\def\paperID{5044}
\def\confName{CVPR}
\def\confYear{2024}

\def\authorBlock{
    Haoxin Chen  \qquad
    Yong Zhang\footnotemark[2] \qquad
    Xiaodong Cun \qquad
    Menghan Xia \\
    Xintao Wang \qquad
    Chao Weng \qquad
    Ying Shan \\ \\
    Tencent AI Lab \\
    Homepage: \url{https://ailab-cvc.github.io/videocrafter} \\
    Github: \url{https://github.com/AILab-CVC/VideoCrafter} \\
    Discord: \url{https://discord.gg/RQENrunu92} 
}

\newif\ifreview 
\newif\ifarxiv 
\newif\ifcamera \newcommand{\cameraready}{\cameratrue}
\newif\ifrebuttal 

\cameraready 

\newcommand{\animation}{0}

\pdfoutput=1
\documentclass[10pt,twocolumn,letterpaper]{article}
\ifreview \usepackage[review]{cvpr} \fi
\ifarxiv \usepackage[pagenumbers]{cvpr} \fi
\ifrebuttal \usepackage[rebuttal]{cvpr} \fi
\ifcamera \usepackage{cvpr} \fi

\usepackage{graphicx, animate}
\usepackage{amsmath}	
\usepackage{amssymb}	
\usepackage{booktabs}
\usepackage{times}
\usepackage{microtype}
\usepackage{epsfig}
\usepackage[table,xcdraw,dvipsnames]{xcolor}
\usepackage{caption}
\usepackage{float}
\usepackage{placeins}
\usepackage{color, colortbl}
\usepackage{stfloats}
\usepackage{enumitem}
\usepackage{tabularx}
\usepackage{xstring}
\usepackage{multirow}
\usepackage{xspace}
\usepackage{url}
\usepackage{subcaption}
\usepackage{xcolor}
\usepackage[hang,flushmargin]{footmisc}

\ifcamera \usepackage[accsupp]{axessibility} \fi

% ...

\ifarxiv  \fi

\newcommand{\R}[1]{{%
    \textbf{%
        \ifstrequal{#1}{1}{\textcolor{red}{R#1}}{%
        \ifstrequal{#1}{2}{\textcolor{blue}{R#1}}{%
        \ifstrequal{#1}{3}{\textcolor{magenta}{R#1}}{%
        \ifstrequal{#1}{4}{\textcolor{teal}{R#1}}{%
                           \textcolor{cyan}{R#1}%
        }}}}%
    }%
}}

\usepackage{xr-hyper}

\makeatletter
\newcommand*{\addFileDependency}[1]{
  \typeout{(#1)}
  \@addtofilelist{#1}
  \IfFileExists{#1}{}{\typeout{No file #1.}}
}

\makeatother

\definecolor{cvprblue}{rgb}{0.21,0.49,0.74}
\usepackage[pagebackref,breaklinks,colorlinks,citecolor=cvprblue]{hyperref}
\usepackage[capitalize]{cleveref}
\crefname{section}{Sec.}{Secs.}
\crefname{table}{Table}{Tables}
\crefname{figure}{Fig.}{Figs.}

\begin{document}

\title{VideoCrafter2: Overcoming Data Limitations for High-Quality  \\ Video Diffusion Models}
\author{\authorBlock}


\if \animation 1

\twocolumn[{
\maketitle
\begin{center}
    \captionsetup{type=figure}
    \includegraphics[width=0.33\linewidth]{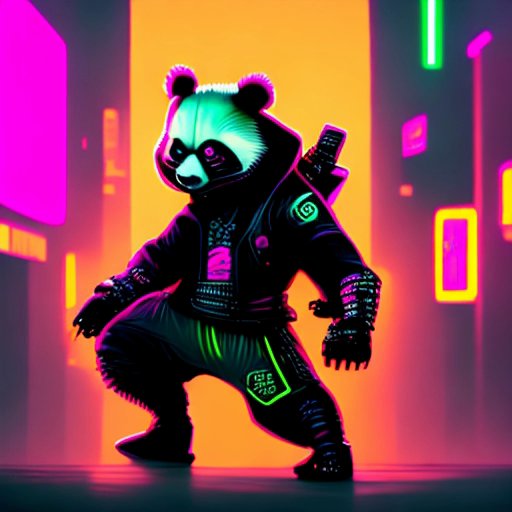}
    \includegraphics[width=0.33\linewidth]{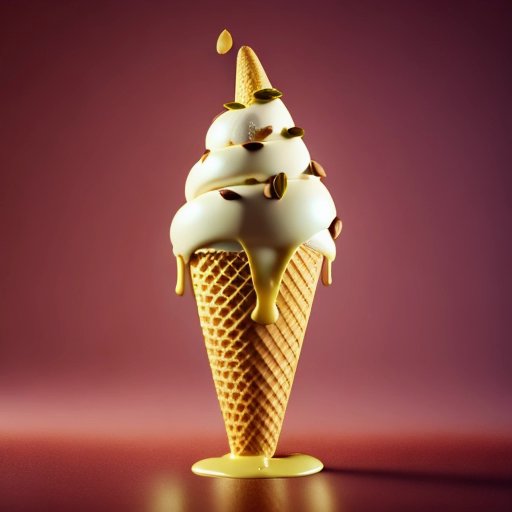} 
    \includegraphics[width=0.33\linewidth]{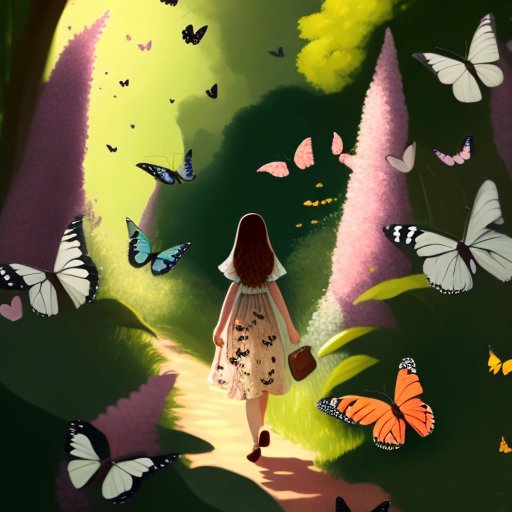}  \\
    {\shortstack[l]{\small \textit{In cyberpunk, neonpunk style,} \\ \small \textit{Kung Fu Panda, jump and kick.}}}  
    \hspace{0.4cm}
    {\shortstack[l]{\small \textit{Cinematic photo melting pistachio ice }\\ \small \textit{ cream dripping down the cone.} \\ \small \textit{ 35mm photograph, film, bokeh}}}   
     \hspace{0.4cm}
    {\shortstack[l]{\small \textit{Large motion, surrounded by butterflies, }\\ \small \textit{ a girl walks through a lush garden.}} }  
    \captionof{figure}{Give a text prompt, our method can generate a video with high visual quality and accurate text-video alignment.  Note that it is trained with only low-quality videos and high-quality images. No high-quality videos are required. \textit{Best viewed with Acrobat Reader. Click the images to play the video clips.}}
     \label{fig:teaser}
\end{center}
}]

\else 

\twocolumn[{
\maketitle
\begin{center}
    \captionsetup{type=figure}
    \animategraphics[width=0.33\linewidth]{8}{fig/sota/0263_}{0}{15}
    \animategraphics[width=0.33\linewidth]{8}{fig/sota/0023_}{0}{15}
    \animategraphics[width=0.33\linewidth]{8}{fig/sota/0436_}{0}{15} \\
    {\shortstack[l]{\small \textit{In cyberpunk, neonpunk style,} \\ 
    \hspace{0.0cm}
    \small \textit{Kung Fu Panda, jump and kick.}}}  
    \hspace{1.2cm}
    {\shortstack[l]{\small \textit{Cinematic photo melting pistachio ice }\\ \small \textit{ cream dripping down the cone.} \\ \small \textit{ 35mm photograph, film, bokeh.}}}   
     \hspace{1cm}
    {\shortstack[l]{\small \textit{Large motion, surrounded by butterflies, }\\ \small \textit{ a girl walks through a lush garden.}} 
    \hspace{0.0cm}
    }  
    \captionof{figure}{Give a text prompt, our method can generate a video with high visual quality and accurate text-video alignment.  Note that it is trained with only low-quality videos and high-quality images. No high-quality videos are required.  \textit{Best viewed with Acrobat Reader. Click the images to play the video clips.}}
    \label{fig:teaser}
\end{center}
}]

\fi

\renewcommand{\thefootnote}{\fnsymbol{footnote}}
\footnotetext[2]{~Corresponding author.}

\begin{abstract}

Text-to-video generation aims to produce a video based on a given prompt.
Recently, several commercial video models have been able to generate plausible videos with minimal noise, excellent details, and high aesthetic scores.
However, these models rely on large-scale, well-filtered, high-quality videos that are not accessible to the community.
Many existing research works, which train models using the low-quality WebVid-10M dataset, struggle to generate high-quality videos because the models are optimized to fit WebVid-10M. 
In this work, we explore the training scheme of video models extended from Stable Diffusion and investigate the feasibility of leveraging low-quality videos and synthesized high-quality images to obtain a high-quality video model.
We first analyze the connection between the spatial and temporal modules of video models and the distribution shift to low-quality videos. We observe that full training of all modules results in a stronger coupling between spatial and temporal modules than only training temporal modules. 
Based on this stronger coupling, we shift the distribution to higher quality without motion degradation by finetuning spatial modules with high-quality images, resulting in a generic high-quality video model.
Evaluations are conducted to demonstrate the superiority of the proposed method, particularly in picture quality, motion, and concept composition.

\end{abstract}
\section{Introduction}
\label{sec:intro}
Benefiting from the development of diffusion models~\cite{ho2020denoising,song2021score}, video generation has achieved breakthroughs, particularly in basic text-to-video (T2V) generation models.
Most existing methods~\cite{ho2022imagenvideo,hu2022make,he2022latent,zhou2022magicvideo,wang2023modelscope,blattmann2023align} follow a logic to obtain video models, \textit{i.e.,} extending a text-to-image (T2I) backbone to a video model by adding temporal modules and then training it with videos.
Several models train video models from scratch, while most start from a pre-trained T2I model, typically Stable Diffusion (SD)~\cite{rombach2022high}.
Models can also be categorized into two groups based on the space modeled by diffusion models, \textit{i.e.,} pixel-space models~\cite{ho2022imagenvideo,hu2022make, zhang2023show1} and latent-space models~\cite{he2022latent,zhou2022magicvideo,wang2023modelscope,blattmann2023align}.
The latter is the dominant approach.
Picture quality, motion consistency, and concept composition are essential dimensions for evaluating a video model.
Picture quality refers to aspects such as sharpness, noise, distortion, aesthetic score, and more.
Motion consistency refers to the appearance consistency between frames and motion smoothness.
Concept composition represents the ability to combine different concepts that might not appear simultaneously in real videos.

Recently, a few commercial startups have released their T2V models~\cite{gen2,pikalab,moonvalley,genmo} that can produce plausible videos with minimal noise, excellent details, and high aesthetic scores. However, they are trained on a large-scale and well-filtered high-quality video dataset, which is not accessible to the community and academia. Collecting millions of high-quality videos is challenging due to copyright restrictions and post-filtering processing. Though there are a few open-source video datasets collected from the Internet for video understanding, such as HowTo100M~\cite{miech19howto100m}, HD-VILA-100M~\cite{xue2022hdvila}, and InterVid~\cite{wang2023internvid}, there exist many issues for video generation, \textit{e.g.,} poor picture quality and caption, multiple clips in one video, and static frames or slides. WebVid-10M~\cite{bain2021frozen} is the most widely used dataset to train video generation models in academia. The clips are well-segmented, and the diversity is good. However, the picture quality is unsatisfactory, and most videos have a resolution of about 320p. The lack of high-quality datasets poses a significant obstacle to training high-quality video models in academia.

In this work, we target a quite challenging problem, \textit{i.e.,} training high-quality video models without using high-quality videos.  
We dive into the training process of SD-based video models to analyze the connection between spatial and temporal modules under different training strategies and investigate the distribution shift to low-quality videos.  
We make a meaningful observation that the full training of all modules results in a stronger coupling between appearance and motion than just training temporal modules. 
The full training can achieve more natural motion and tolerate more subsequent modification of spatial modules, which is the key to improving the quality of generated videos.   
Based on the observation of the connection, we propose a method to overcome the data limitation by disentangling motion from appearance at the data level. 
Specifically, instead of high-quality videos, we exploit low-quality videos to guarantee motion consistency and use high-quality images to ensure picture quality and concept composition ability. 
Benefiting from the successful T2I models such as SDXL and Midjourney, it is convenient to obtain a large set of images with high-resolution and complex concept composition.
Following the guidelines from the analysis, we design a pipeline to fully train a video model extended from SD. 
Then, by exploring different ways of modifying the spatial and temporal modules of the fully trained model with synthesized images, we identify that finetuning spatial weights only is better than other ways, and directly finetuning is better than LORA~\cite{hu2021lora}.
Fig.~\ref{fig:teaser} shows visual examples generated by our method.

Our main contributions are summarized as follows:
\begin{itemize}
    \item We propose a method to overcome the data for training high-quality video models by disentangling motion from appearance at the data level.
    \item We investigate the connection between spatial and temporal modules, and the distribution shift. We identify the keys to obtain a high-quality video model.
    \item We design an effective pipeline based on the observations, \textit{i.e.,} obtaining a fully trained video model first and tuning the spatial modules with synthesized high-quality images.
\end{itemize}

\section{Related Work}
\label{sec:related}
The evolution of video generation techniques goes along with the development of generative models. Generative adversarial networks~\cite{creswell2018generative} and variational auto-encoders~\cite{doersch2016tutorial} are the commonly used backbones in early research of video generation, \textit{e.g.}, VGAN~\cite{vondrick2016generating}, TGAN~\cite{saito2017temporal}, MoCoGAN~\cite{tulyakov2018mocogan}, GODIA~\cite{wu2021godiva}, StyleGAN-V~\cite{skorokhodov2022stylegan}, and MCVD~\cite{voleti2022mcvd}. Then, since transformers have been successfully applied in various fields, they are also introduced for video synthesis, \textit{e.g.}, CogVideo~\cite{hong2022cogvideo}, VideoGPT~\cite{yan2021videogpt}, NUVA-infinity~\cite{wu2022nuwa}, TATS~\cite{ge2022long}, MAGVIT~\cite{yu2023magvit}, Phenaki~\cite{villegas2022phenaki}.

Recently, diffusion models (DMs)~\cite{sohl2015deep,ho2020denoising,song2021score} have been a famous star in generative models, especially in text-to-image (T2I) generation~\cite{ramesh2022hierarchical,nichol2022glide,saharia2022photorealistic,gu2022vector,rombach2022high,balaji2022ediffi,he2023scalecrafter,zhang2023real,podell2023sdxl}. For video generation, Video Diffusion Models (VDMs) are proposed to model the distribution of videos. VDM~\cite{ho2022video} is the first to utilize a space-time factorized U-Net to model videos in pixel space for unconditional video generation. It uses an image-video joint training strategy to avoid concept forgetting. Imagen Video~\cite{ho2022imagenvideo} and Make-a-Video~\cite{hu2022make} are two cascade models that target text-to-video generation in pixel space. Show-1~\cite{zhang2023show1} is another cascade model that uses IF~\cite{IF2023} as the base model and LDM extended video model for super-resolution. Subsequently, LVDM~\cite{he2022latent,chen2023videocrafter1} and MagicVideo~\cite{zhou2022magicvideo} propose to extend LDM~\cite{rombach2022high} to model videos in the latent space of an auto-encoder. Many other methods use the same paradigm, including ModelScope~\cite{wang2023modelscope}, Align Your Latent~\cite{blattmann2023align}, Hotshot-XL~\cite{hotshotxl}, LAVIE~\cite{wang2023lavie}, PYOCO~\cite{ge2023preserve}, VideoFactory~\cite{wang2023videofactory}, VPDM~\cite{yu2023video}, VIDM~\cite{mei2023vidm}, and Latent-Shift~\cite{an2023latent}. Besides text-to-video generation, a few methods, such as~\cite{xing2023dynamicrafter, chen2023seine, zhang2023i2vgenxl}, attempt to generate videos from a given image and a prompt as condition. 

Several startups release their text-to-video generation services, \textit{e.g.,} Gen-2~\cite{gen2}, Pika Labs~\cite{pikalab}, Moonvalley~\cite{moonvalley}, and Genmo~\cite{genmo}. Their models can generate plausible results with minimal noise, excellent details, and high aesthetic scores. However, those methods are trained with a large-scale well-filtered high-quality video dataset that is not accessible to researchers. The video models are also not available, leading to the slow development of downstream tasks to a certain extent. The most widely used video dataset is WebVid-10M, a large-scale dataset of short videos with textual descriptions sourced from stock footage sites. The videos are diverse and rich in their content, and each video is well-segmented, however, the picture quality is unsatisfactory and most videos are 320p. Training a high-quality video model under the data limitation is quite challenging.

AnimateDiff~\cite{guo2023animatediff} finds that combining temporal modules from a video model trained on WebVid-10M and a LORA SD model can improve the picture quality of the generated videos. 
However, this is not a generic model and does not always work. There are a few severe issues. 
First, the temporal modules can only be combined with a few selected LORA models, which makes it not a generic model. 
Second, since each LORA model is a personalized model, the composed video model might suffer from degraded concept composition if the LORA model trained with limited data. 
Third, the motion quality degenerates when the modules do not match the LORA model well. 
Unlike AnimateDiff, we analyze the connection between spatial and temporal modules  instead of direct combination, and design a pipeline to train a generic high-quality video model without high-quality video by disentangling appearance and motion at the data level. 
\section{Method}
\label{sec:method}
We propose an effective method to overcome the data limitation for training high-quality video diffusion models. We first analyze the connection between spatial and temporal modules of SD-based video models under different training strategies. Based on the observations, we then develop a pipeline to train high-quality video models with just low-quality videos and high-quality images, \textit{i.e.,} disentangling appearance from motion at the data level.

\subsection{Spatial-temporal Connection Analyses} \label{sec:connection}
\paragraph{Base T2V model.} 
To leverage the prior in SD trained on a large-scale image dataset, most text-to-video diffusion models inflate the SD model to a video model by adding temporal modules, including Align Your Latent~\cite{blattmann2023align}, AnimateDiff~\cite{guo2023animatediff}, LVDM~\cite{he2022latent}, Magic Video~\cite{zhou2022magicvideo}, ModelScopeT2V~\cite{wang2023modelscope}, and LAVIE~\cite{wang2023lavie}. They follow VDM~\cite{ho2022video} to use a particular type of 3D-UNet that is factorized over space and time.

These models can be categorized into two groups according to their training strategies. One is to use videos to learn both the spatial and temporal modules with the SD weights as initialization, called \textit{full training}. The other is to train temporal modules with spatial ones fixed, called \textit{partial training}. Align Your Latent and AnimateDiff belong to the first group, while other T2V models belong to the other group.

Though these SD-based T2V models have similar architectures, they are trained under different training settings. We use one typical model to investigate the connection between spatial and temporal modules under the two training strategies. We follow the architecture of the open-source VideoCrafter1~\cite{chen2023videocrafter1} with FPS (frames per second) condition. We also incorporate the temporal convolution in ModelScopeT2V~\cite{wang2023modelscope} to improve temporal consistency.

\paragraph{Parameter Perturbation for Full and Partial Training.}
We apply the two training strategies to the same architecture using the same data. The model is initialized from pretrained SD weights. WebVid-10M~\cite{bain2021frozen} is exploited as the training data. To avoid concept forgetting, LAION-COCO 600M~\cite{LAION-COCO} is also used for video and image joint training. The resolution is $512\times320$. For simplicity, the fully trained video model is denoted as $M_F (\theta_T, \theta_S)$, while the partially trained one is denoted as $M_P(\theta_T, \theta_S^0)$, where $\theta_T$ and $\theta_S$ are the learned parameters of the temporal and spatial modules, respectively. $\theta_S^0$ are the original spatial parameters of SD.

To evaluate the connection strength between spatial and temporal modules, we perturb the parameters of the specified modules by using another high-quality image dataset $\mathcal{D}_I$ for finetuning. The image data is JDB~\cite{pan2023journeydb} that consists of synthesized images from Midjornery~\cite{Midjourney}. As the JDB has 4 million images, we use LORA~\cite{hu2021lora} for finetuning.

\if \animation 1

\begin{figure}[t]
  \centering
  \setlength{\tabcolsep}{0.1pt} 
  \begin{tabular}{>{\centering\arraybackslash}m{0.5cm} >{\centering\arraybackslash}m{3.8cm} >{\centering\arraybackslash}m{3.8cm} }
     & \textit{$\text{iter}=1K$ } & $\text{iter}=10K$ \\
    \rotatebox{90}{ \small P-Spa-LORA  &
    \includegraphics[width=\linewidth]{fig/comp_lora_spatial/fix-1k/0012_0.jpg} &
    \includegraphics[width=\linewidth]{fig/comp_lora_spatial/fix-10k/0012_0.jpg}  \\
    \rotatebox{90}{ \small F-Spa-LORA } &
    \includegraphics[width=\linewidth]{fig/comp_lora_spatial/full-1k/0012_0.jpg} &
    \includegraphics[width=\linewidth]{fig/comp_lora_spatial/full-10k/0012_0.jpg}  \\
  \end{tabular}
  \caption{Perturbing spatial modules using LORA. \textit{Best viewed with Acrobat Reader. Click the images to play the video clips.}}
\end{figure}

\else

\begin{figure}[t]
  \centering
  \setlength{\tabcolsep}{0.1pt} 
  \begin{tabular}{>{\centering\arraybackslash}m{0.5cm} >{\centering\arraybackslash}m{3.8cm} >{\centering\arraybackslash}m{3.8cm} }
     & \textit{$\text{iter}=1K$ } & $\text{iter}=10K$ \\
    \rotatebox{90}{\small P-Spa-LORA } &
    \animategraphics[width=\linewidth]{8}{fig/comp_lora_spatial/fix_1k/0012_}{0}{15}&
    \animategraphics[width=\linewidth]{8}{fig/comp_lora_spatial/fix_10k/0012_}{0}{15} \\
    \rotatebox{90}{\small F-Spa-LORA } &
    \animategraphics[width=\linewidth]{8}{fig/comp_lora_spatial/all_1k/0012_}{0}{15} &
    \animategraphics[width=\linewidth]{8}{fig/comp_lora_spatial/all_10k/0012_}{0}{15}  \\
    \multicolumn{3}{c}{ \small \textit{A professional dancer gracefully performs ballet on stage.}}
  \end{tabular}
  \vspace{-3mm}
  \caption{Perturbing spatial modules using LORA. \textit{Best viewed with Acrobat Reader. Click the images to play the video clips.}}
  \label{fig:connection-spatial}
  \vspace{-3mm}
\end{figure}

\fi

\if \animation 1

\begin{figure}[t]
  \centering
  \setlength{\tabcolsep}{0.1pt} 
  \begin{tabular}{>{\centering\arraybackslash}m{0.5cm} >{\centering\arraybackslash}m{3.8cm} >{\centering\arraybackslash}m{3.8cm} }
     & \textit{Robot dancing in times square } & \textit{Beautiful pink rose background. blooming rose flower rotation, close-up.} \\
    \rotatebox{90}{\small P-Temp-LORA } &
    \includegraphics[width=\linewidth]{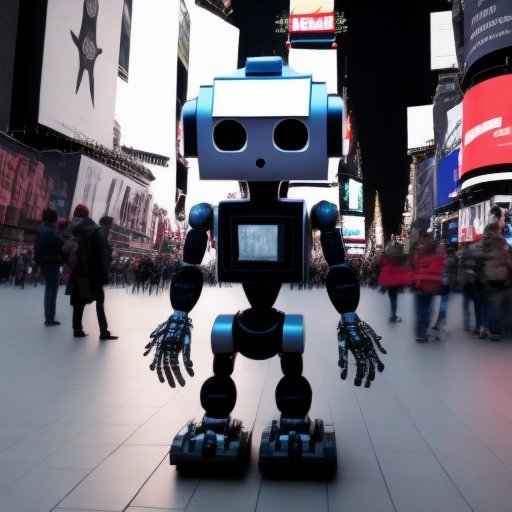} &
    \includegraphics[width=\linewidth]{{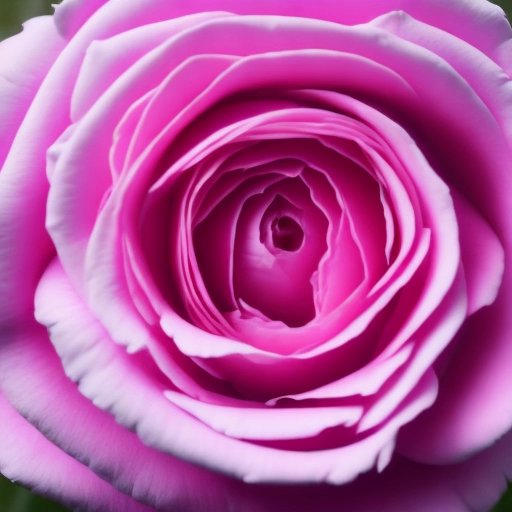}  \\
    \rotatebox{90}{\small F-Temp-LORA } &
    \includegraphics[width=\linewidth]{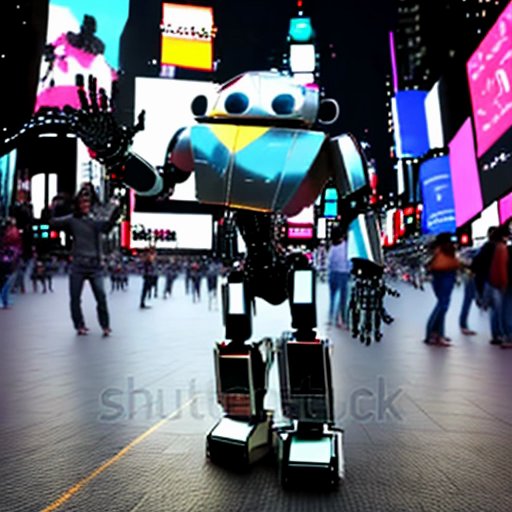} &
    \includegraphics[width=\linewidth]{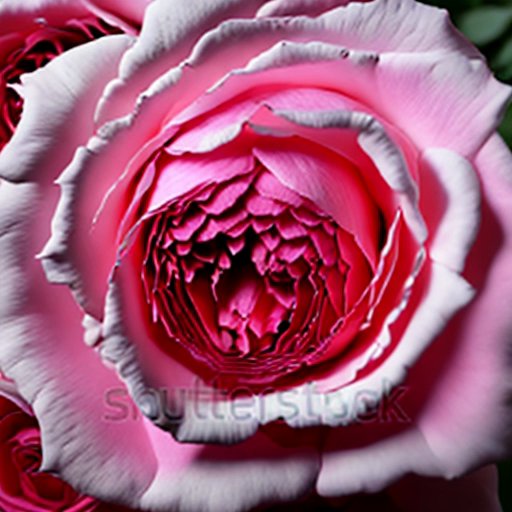}  \\
  \end{tabular}
  \caption{Perturbing temporal modules using LORA.  \textit{Best viewed with Acrobat Reader. Click the images to play the video clips.}}
  \label{fig:connection-temp}
\end{figure}

\else

\begin{figure}[t]
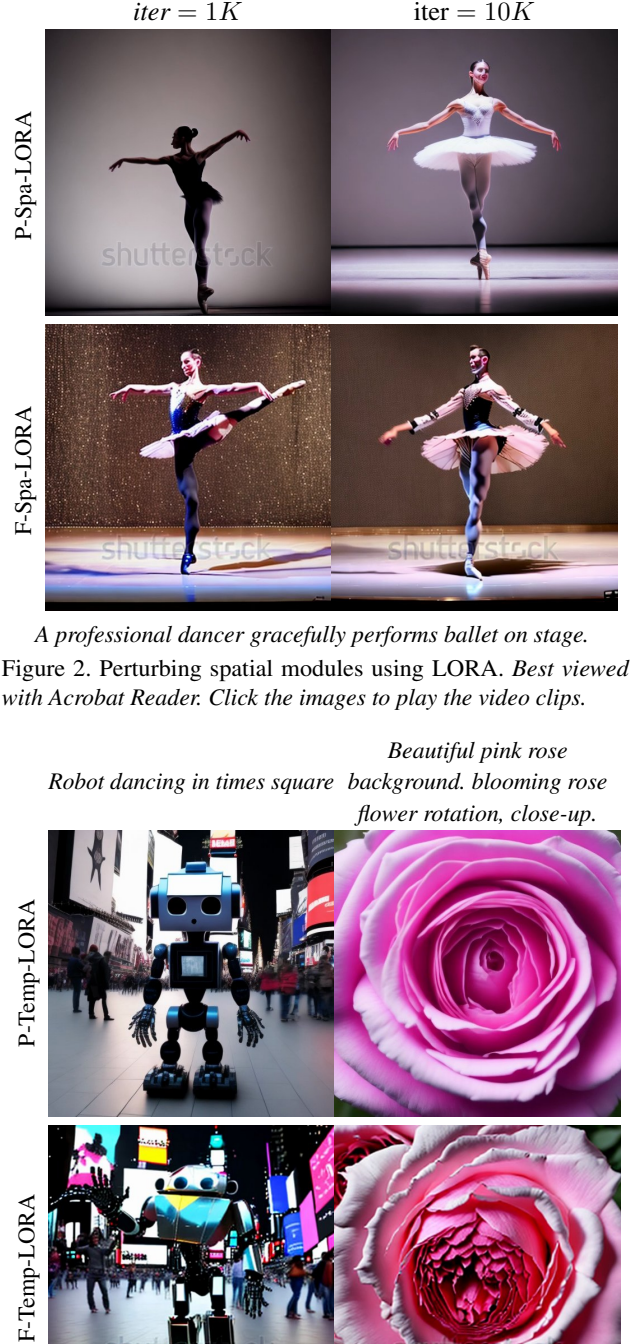

  \centering
  \setlength{\tabcolsep}{0.1pt} 
  \begin{tabular}{>{\centering\arraybackslash}m{0.5cm} >{\centering\arraybackslash}m{3.8cm} >{\centering\arraybackslash}m{3.8cm} }
     & {\small \textit{Robot dancing in times square }} & { \small \textit{Beautiful pink rose background. blooming rose flower rotation, close-up.}} \\
    \rotatebox{90}{ \small P-Temp-LORA } &
    \animategraphics[width=\linewidth]{8}{fig/comp_lora_temp/fix_10k/0015_}{0}{15}&
    \animategraphics[width=\linewidth]{8}{fig/comp_lora_temp/fix_10k/0025_}{0}{15} \\
    \rotatebox{90}{ \small F-Temp-LORA } &
    \animategraphics[width=\linewidth]{8}{fig/comp_lora_temp/all_10k/0015_}{0}{15} &
    \animategraphics[width=\linewidth]{8}{fig/comp_lora_temp/all_10k/0025_}{0}{15}  \\
  \end{tabular}
  \caption{Perturbing temporal modules using LORA.   \textit{Best viewed with Acrobat Reader. Click the images to play the video clips.}}
  \label{fig:connection-temp}
\end{figure}

\fi
\textit{Spatial Perturbation.} We first perturb the spatial parameters of the two video models using the image dataset. The temporal parameters are frozen. The perturbation process of the fully trained base model $M_F$ can be denoted as:
\begin{equation}
    M_F^{'} (\theta_T, \theta_S + \Delta_{\theta_S}) \leftarrow \text{PERTB}^{\text{LORA}}_{\theta_S} (M_F (\theta_T, \theta_S), \mathcal{D}_I), \nonumber
\end{equation}
where $\text{PERTB}^{\text{LORA}}_{\theta_S}$ denote finetuning $M_F$ with respect to $\theta_S$ on the image dataset $\mathcal{D}_I$ using LORA. 
$\Delta_{\theta_S}$ represents the parameters of the LORA branch. 
Similarly, we can obtain the perturbed model of the partially trained video model: 
\begin{equation}
    M_P^{'} (\theta_T, \theta_S^0 + \Delta_{\theta_S}) \leftarrow \text{PERTB}^{\text{LORA}}_{\theta_S} (M_P (\theta_T, \theta_S^0), \mathcal{D}_I). \nonumber
\end{equation}

For easy understanding, we also use the name `F-Spa-LORA' to denote model $M_F^{'}$ and `P-Spa-LORA' for $M_P^{'}$. `F' denotes the fully trained base model while `P' stands for the partially trained model. `Spa' and `Temp' mean finetuning spatial and temporal modules, respectively. `LORA' represents using LORA for finetuning, while `DIR' means direct finetuning without LORA. For example, `F-Spatial-LORA' represents perturbing spatial modules of the fully trained T2V model using LORA.

Comparing the synthesized videos of the two resulting models, we have the following observations. First, the motion quality of F-Spa-LORA is more stable than that of P-Spa-LORA (see user study in Table~\ref{tab:pertb-user}). The motion of P-Spa-LORA becomes worse quickly during the finetuning process. The more finetuning steps, the video tends to be more still with local flicker (see Fig.~\ref{fig:connection-spatial}). While the motion of F-Spa-LORA slightly degenerates compared to the fully trained base model. Second, P-Spa-LORA achieves much better visual quality than F-Spa-LORA (see Fig.~\ref{fig:connection-spatial}). The picture quality and aesthetic score of F-Spa-LORA are greatly improved compared to the partially trained base model (see Table~\ref{tab:pertb-vis}). Surprisingly, the watermark is also removed. While F-Spa-LORA obtains a slight improvement in picture quality and aesthetic score, the generated videos are still noisy.

From the two observations, we can conclude that the coupling strength between spatial and temporal modules of the fully trained model is stronger than that of the partially trained model. Because the spatial-temporal coupling of the partially trained model can be easily broken, leading to quick motion degeneration and picture quality shift. A stronger connection can tolerate parameter perturbation more than a weak one. Our observation can be used to explain the quality improvement and motion degeneration of AnimateDiff. AnimateDiff is not a generic model and only works for selected personalized SD models. The reason is that its motion modules are obtained with the partially training strategy, and they cannot tolerate large parameter perturbations. When the personalized model does not match the temporal modules, both picture and motion quality will degenerate.

\textit{Temporal Perturbation.} 
The partially trained model has only the temporal modules updated, but the picture quality is shifted to the quality of WebVid-10M. Hence, the temporal modules take responsibility for not only the motion but also the picture quality. We perturb the temporal modules while fixing the spatial modules with the image dataset. The perturbation processes can be denoted as:
\begin{align}
    M_F^{''} (\theta_T + \Delta_{\theta_T}, \theta_S ) &\leftarrow \text{PERTB}^{\text{LORA}}_{\theta_T} (M_F (\theta_T, \theta_S), \mathcal{D}_I), \nonumber \\
    M_P^{''} (\theta_T+ \Delta_{\theta_T}, \theta_S^0 ) &\leftarrow \text{PERTB}^{\text{LORA}}_{\theta_T} (M_P (\theta_T, \theta_S^0), \mathcal{D}_I).\nonumber
\end{align}
We observe that the picture quality of P-Temp-LORA ($M_P^{''}$) is better than F-Temp-LORA ($M_F^{''}$). However, the foreground and background of the videos are more shaky, \textit{i.e.,} the temporal consistency becomes worse (see Fig.~\ref{fig:connection-temp}).  The picture of F-Temp-LORA is improved, but the watermark is still there. Its motion is close to the base model and much better than P-Temp-LORA (see Table~\ref{tab:pertb-user}). Those observations also support the conclusion obtained from spatial perturbation.

\if \animation 1

\begin{figure*}[ht]
  \centering
  \setlength{\tabcolsep}{0.1pt} 
  \begin{tabular}{>{\centering\arraybackslash}m{4.2cm} >{\centering\arraybackslash}m{4.2cm} >{\centering\arraybackslash}m{4.2cm} >{\centering\arraybackslash}m{4.2cm} }
     {\small \textit{F-Spa\&Temp-LORA}} & {\small  \textit{F-Temp-DIR} } & {\small \textit{F-Spa-DIR} } & {\small  \textit{F-Spa\&Temp-DIR} }\\
    \includegraphics[width=\linewidth]{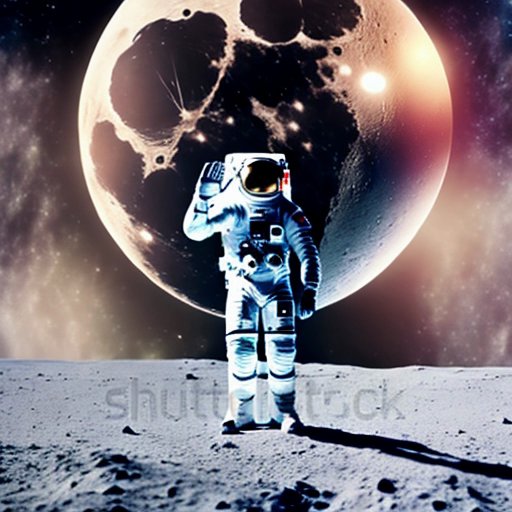} &
    \includegraphics[width=\linewidth]{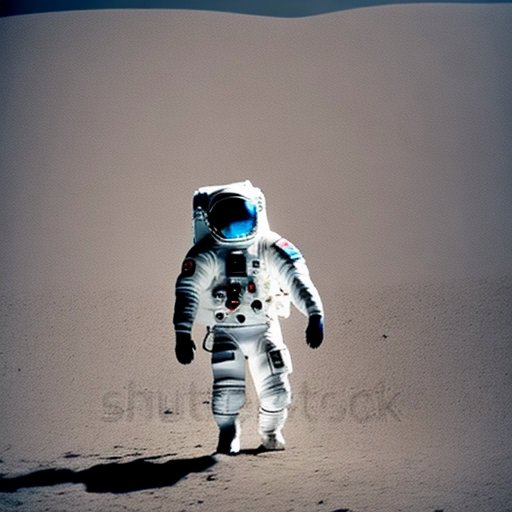} &
    \includegraphics[width=\linewidth]{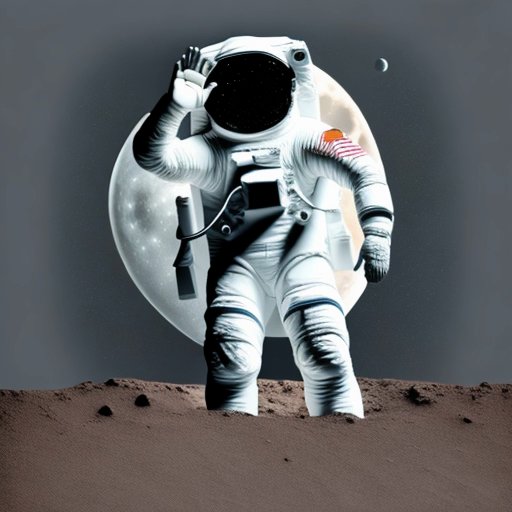} &
    \includegraphics[width=\linewidth]{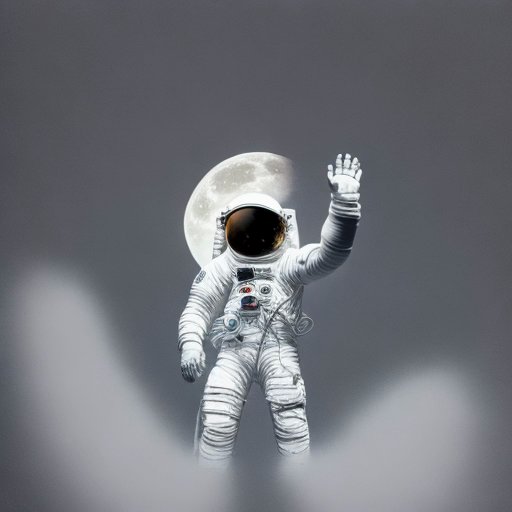} \\
  \end{tabular}
  \caption{Module selection based on the fully trained T2V model. \textit{Best viewed with Acrobat Reader. Click the images to play the video clips.}}
  \label{fig:disentangle-module}
\end{figure*}

\else

\begin{figure*}[ht]
  \centering
  \setlength{\tabcolsep}{0.1pt} 
  \begin{tabular}{>{\centering\arraybackslash}m{4.2cm} >{\centering\arraybackslash}m{4.2cm} >{\centering\arraybackslash}m{4.2cm} >{\centering\arraybackslash}m{4.2cm} }
     {\small {F-Spa\&Temp-LORA}} & {\small  {F-Temp-DIR} } & {\small {F-Spa-DIR} } & {\small  {F-Spa\&Temp-DIR} }\\
    \animategraphics[width=\linewidth]{8}{fig/comp_dir_part/spatial_temp_lora/0036_}{0}{15}&
    \animategraphics[width=\linewidth]{8}{fig/comp_dir_part/temp/0036_}{0}{15}&
    \animategraphics[width=\linewidth]{8}{fig/comp_dir_part/spatial/0036_}{0}{15}&
    \animategraphics[width=\linewidth]{8}{fig/comp_dir_part/spatial_temp/0036_}{0}{15} \\
     \multicolumn{4}{c}{\textit{An astronaut is waving his hands on the moon.} }
  \end{tabular}
  \vspace{-2mm}
  \caption{Module selection based on the fully trained T2V model. \textit{Best viewed with Acrobat Reader. Click the images to play the video clips.}}  \label{fig:disentangle-module}
  \vspace{-2mm}
\end{figure*}

\fi

\subsection{Data-level Disentanglement of Appearance and Motion} \label{sec:disentanglement}
Since obtaining a large-scale, high-quality video dataset with high diversity is challenging due to copyright issues, we explore the possibility of training a high-quality video model without using high-quality videos. 
We have access to low-quality videos such as WebVid-10M and high-quality images such as JDB. 
We propose to disentangle motion from appearance at the data level, \textit{i.e.,} learning motion from low-quality videos while learning picture quality and aesthetics from high-quality images.  
We can first train a video model with videos and then fine-tune the video model with images. 
\textit{The keys lie in how to train a video model and how to fine-tune it with images.}

According to the study of the connection between spatial and temporal modules, 
a fully trained model is more suitable for the subsequent finetuning with high-quality images. This is because the strong spatial-temporal coupling can tolerate the parameter perturbation for both spatial and temporal modules without obvious motion degeneration.   

\if \animation 1

\begin{figure}[t]
  \centering
  \setlength{\tabcolsep}{0.1pt} 
  \begin{tabular}{ >{\centering\arraybackslash}m{3.8cm} >{\centering\arraybackslash}m{3.8cm} }
    & \textit{anime illustration of a blue pig, riding a scooter near a lake, with the sun in the sky} \\
    \includegraphics[width=\linewidth]{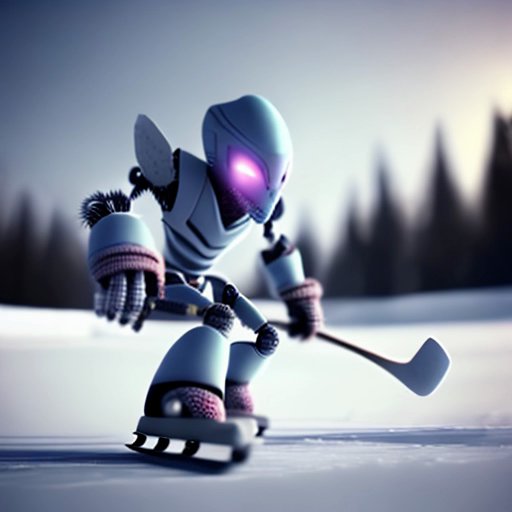} &
    \includegraphics[width=\linewidth]{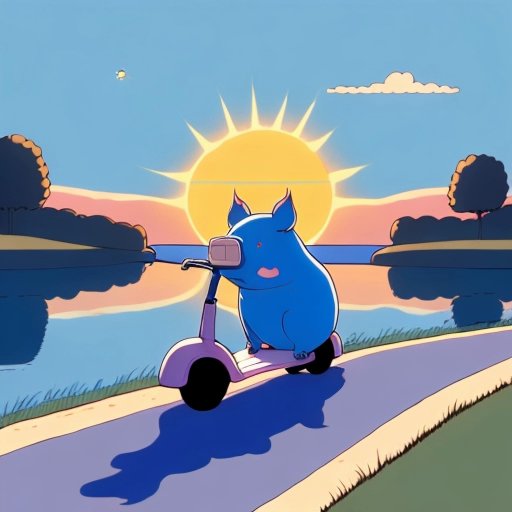}  \\
  \end{tabular}
  \caption{Influence of image data on concept composition.  \textit{Best viewed with Acrobat Reader. Click the images to play the video clips.}}
  \label{fig:data}
\end{figure}

\else

\begin{figure}[t]
  \centering
  \setlength{\tabcolsep}{0.1pt} 
  \begin{tabular}{ >{\centering\arraybackslash}m{4cm} >{\centering\arraybackslash}m{4cm} }
  \multicolumn{2}{p{7cm}}{{\small \textit{Anime illustration of a blue pig, riding a scooter near a lake, with the sun in the sky}}}
      \\
    \animategraphics[width=\linewidth]{8}{fig/comp_data/laion/0041_}{0}{15} &
    \animategraphics[width=\linewidth]{8}{fig/comp_data/jdb/0041_}{0}{15}  \\
     {\small F-Spa-DIR-LAION} & {\small F-Spa-DIR }
  \end{tabular}
  \caption{Influence of image data on concept composition. 
  `F-Spa-DIR-LAION' uses the LAION aesthetics V2 as the image data while  `F-Spa-DIR' uses JDB. 
  \textit{Best viewed with Acrobat Reader. Click the images to play the video clips.}}
  \label{fig:data}
\end{figure}
\fi

Next, we need to investigate how to fine-tune the base model with images. 
In both spatial and temporal perturbation (Sec.~\ref{sec:connection}), the picture quality can be improved but not very significantly. 
To obtain a greater quality improvement, we evaluate two strategies. 
One is to involve more parameters, \textit{i.e.,} finetuning both spatial and temporal modules with images. 
The other is to change the finetuning method, \textit{i.e.,} using direct finetuning without LORA. 
We can evaluate the following four cases:
\begin{align}
    M_F^{A} (\theta_T + \Delta_{\theta_T}, \theta_S + \Delta_{\theta_S} ) &\leftarrow \text{PERTB}^{\text{LORA}}_{\theta_T, \theta_S} (M_F (\theta_T, \theta_S), \mathcal{D}_I), \nonumber \\
    M_F^{B} (\theta_T, \theta_S + \Delta_{\theta_S} ) &\leftarrow \text{PERTB}_{ \theta_S} (M_F (\theta_T, \theta_S), \mathcal{D}_I), \nonumber \\
    M_F^{C} (\theta_T + \Delta_{\theta_T}, \theta_S ) &\leftarrow \text{PERTB}_{\theta_T} (M_F (\theta_T, \theta_S), \mathcal{D}_I), \nonumber   \\
    M_F^{D} (\theta_T + \Delta_{\theta_T}, \theta_S + \Delta_{\theta_S} ) &\leftarrow \text{PERTB}_{\theta_T, \theta_S} (M_F (\theta_T, \theta_S), \mathcal{D}_I), \nonumber
\end{align}
Where $M_F^{A}$ (F-Spa\&Temp-LORA) is obtained by following the first strategy, $M_F^{B}$, $M_F^{C}$, and $M_F^{D}$ are obtained using the second strategy. $M_F^{B}$ (F-Spa-DIR) and $M_F^{C}$ (F-Temp-DIR) represent directly finetuning the spatial and temporal modules, respectively. $M_F^{D}$ (F-Spa\&Temp-DIR) represents directly finetuning all modules.

Comparing the generated videos of the four models, we have the following observations. First, F-Spa\&Temp-LORA can further improve the picture quality of F-Spa-LORA, but the quality is still unsatisfying. The watermark exists in most generated videos, and the noise is obvious. Second, F-Temp-DIR achieves better picture quality than F-Temp-LORA. It is also better than F-Spa\&Temp-LORA. The watermark is removed or lightened in half of the videos. Third, F-Spa-DIR and F-Spa\&Temp-DIR achieve the best picture quality among the fine-tuned models. However, the motion of F-Spa-DIR is better (see Fig.~\ref{fig:disentangle-module} and Table~\ref{tab:pertb-user}). The foreground and background of F-Spa\&Temp-DIR are flashing in videos generated by $M_F^{D}$, especially local textures.

By exploring the finetuning strategies and different modules, we identify that directly finetuning spatial modules with high-quality images is the best way to improve the picture quality without marginal loss of motion quality. At this point, our data-level disentanglement pipeline can be summarized as follows: fully training a video model with low-quality videos first and then directly finetuning the spatial modules only with high-quality images.

\subsection{Promotion of Concept Composition}
To improve the concept composition ability of video models, we propose to use synthesized images with complex concepts instead of using real images at the partial finetuning stage.
The success of T2I models such as SDXL and Midjornery is built upon large-scale high-quality images. They have the ability to composite concepts that do not appear in the real world. 
Rather than using their training images, we propose transferring their concept composition ability to video models by synthesizing a set of images with complex concepts. 
In this way, we can alleviate the burden of capturing both concept and motion well at the same time.

To validate the effectiveness of synthesized images, we use JDB and LAION-aesthetics V2~\cite{LAION-Aesthetics} as image data for the second finetuning stage. 
LAION-aesthetics V2 consists of web-collected images while JDB contains images synthesized by Midjourney. 
We observe that the model trained with JDB has much better concept composition ability (see Fig.~\ref{fig:data} and Table~\ref{tab:pertb-vis}). 
More results are in the supplementary material.

\section{Experiments}
\label{sec:experiments}
\subsection{Settings}

\paragraph{Data.}
To overcome data limitations, we utilized WebVid-10M~\cite{bain2021frozen} as the source of low-quality video data and JDB~\cite{pan2023journeydb} for high-quality image data.
WebVid-10M is a large-scale, diverse video dataset comprising approximately 10 million text-video pairs. The resolution of most videos is $336 \times 596$, and each video consists of a single shot. During training, we sample from videos at varying frame rates.
JDB is a large-scale image dataset featuring around 4 million high-resolution images from Midjourney, each annotated with a corresponding text prompt.
To prevent concept forgetting during the training of the base T2V model, we also employ LAION-COCO\cite{LAION-COCO}, a dataset comprising 600 million generated high-quality captions for publicly available web images, for both image and video training.

\paragraph{Metrics.}
We exploit EvalCrafter~\cite{liu2023evalcrafter} for quantitative evaluation. EvalCrafter is a benchmark to evaluate text-to-video generation models, which contains around 18 objective metrics for visual quality, content quality, motion quality, and text-caption alignment. It provides about 512 prompts.
The objective metrics are aligned to user opinions from five subjective studies, i.e., motion quality, text-video alignment, temporal consistency, visual quality, and user favor.
The motion quality considers three metrics: action recognition, average flow, amplitude classification score, while temporal consistency considers warping error, semantic consistency, face consistency.
The technical and aesthetic scores in EvalCrafter are adapted from DOVER~\cite{wu2023exploring}.
Besides, we conduct user studies of human preference since there still lacks a comprehensive objective metric to measure motion quality.

\paragraph{Training Details.}
In Sec~\ref{sec:connection}, the two based models share the same architecture, adapted from the open-source VideoCrafter1~\cite{chen2023videocrafter1}, and incorporate temporal convolution from ModelScopeT2V~\cite{wang2023modelscope}. 
The spatial modules are initialized with weights from SD 2.1, and the outputs of the temporal modules are initialized to zeros. 
The training resolution is set at $512 \times 320$. 
For joint image and video training, we utilize the low-quality WebVid-10M and LAION-COCO datasets. 
The models are trained on 32 NVIDIA A100 GPUs for $270K$ iterations with a batch size of 128.  
The learning rate is set at $5\times 10^{-5}$ for all training tasks. 
When employing LORA for the perturbation of temporal or spatial modules, we exclusively use JDB for tuning. The finetuning is conducted on 8 A100 GPUs for $30K$ iterations with a batch size of 256.  Given that the images from JDB have a square resolution, we adjust the finetuning resolution to $512\times 512$. 

\if \animation 1

\begin{figure*}[ht]
  \centering
  \setlength{\tabcolsep}{0pt} 
  \begin{tabular}{>{\centering\arraybackslash}m{0.8cm} >{\centering\arraybackslash}m{4.2cm} >{\centering\arraybackslash}m{4.2cm} >{\centering\arraybackslash}m{4.2cm} >{\centering\arraybackslash}m{4.2cm} }
     & \textit{A bear rummages through a dumpster, searching for food scraps.} & \textit{A group of children build a snowman together.} & \textit{A man cruises through the city on a motorcycle, feeling the adrenaline rush} & \textit{A monkey eating a pizza in central park, GoPro film style} \\
    \rotatebox{90}{Gen-2} &
    \includegraphics[width=\linewidth]{example-image-c} &
    \includegraphics[width=\linewidth]{example-image-c} &
    \includegraphics[width=\linewidth]{example-image-c} &
    \includegraphics[width=\linewidth]{example-image-c} \\
    \rotatebox{90}{Pika Labs} &
    \includegraphics[width=\linewidth]{example-image-c} &
    \includegraphics[width=\linewidth]{example-image-c} &
    \includegraphics[width=\linewidth]{example-image-c} &
    \includegraphics[width=\linewidth]{example-image} \\
    \rotatebox{90}{VideoCrafter1} &
    \includegraphics[width=\linewidth]{example-image-a} &
    \includegraphics[width=\linewidth]{example-image-b} &
    \includegraphics[width=\linewidth]{example-image-c} &
    \includegraphics[width=\linewidth]{example-image} \\
    \rotatebox{90}{Show-1} &
    \includegraphics[width=\linewidth]{example-image-a} &
    \includegraphics[width=\linewidth]{example-image-b} &
    \includegraphics[width=\linewidth]{example-image-c} &
    \includegraphics[width=\linewidth]{example-image} \\
    \rotatebox{90}{AnimeDiff} &
    \includegraphics[width=\linewidth]{example-image-a} &
    \includegraphics[width=\linewidth]{example-image-b} &
    \includegraphics[width=\linewidth]{example-image-c} &
    \includegraphics[width=\linewidth]{example-image} \\
    \rotatebox{90}{Ours} &
    \includegraphics[width=\linewidth]{example-image-a} &
    \includegraphics[width=\linewidth]{example-image-b} &
    \includegraphics[width=\linewidth]{example-image-c} &
    \includegraphics[width=\linewidth]{example-image} \\
  \end{tabular}
  \caption{Comparison of different text-to-video generation models}
  \label{fig:sota}
\end{figure*}

\else

\begin{figure*}[ht]
  \centering
  \setlength{\tabcolsep}{0pt} 
  \begin{tabular}{>{\centering\arraybackslash}m{0.8cm} >{\centering\arraybackslash}m{4.2cm} >{\centering\arraybackslash}m{4.2cm} >{\centering\arraybackslash}m{4.2cm} >{\centering\arraybackslash}m{4.2cm} }
     & \textit{A bear rummages through a dumpster, searching for food scraps} & \textit{A Time 1980 painting of a boy going to school} & \textit{A man cruises through the city on a motorcycle, feeling the adrenaline rush} & \textit{A monkey eating a pizza in central park, GoPro film style} \\
    \rotatebox{90}{\normalsize Gen-2} &
    \animategraphics[width=\linewidth]{8}{fig/comp_sota/gen2/0045_}{0}{15}&
    \animategraphics[width=\linewidth]{8}{fig/comp_sota/gen2/0330_}{0}{15}&
    \animategraphics[width=\linewidth]{8}{fig/comp_sota/gen2/0137_}{0}{15}&
    \animategraphics[width=\linewidth]{8}{fig/comp_sota/gen2/0267_}{0}{15} \\
    \rotatebox{90}{\normalsize Pika Labs} &
    \animategraphics[width=\linewidth]{8}{fig/comp_sota/pika/0045_}{0}{15}&
    \animategraphics[width=\linewidth]{8}{fig/comp_sota/pika/0330_}{0}{15}&
    \animategraphics[width=\linewidth]{8}{fig/comp_sota/pika/0137_}{0}{15}&
    \animategraphics[width=\linewidth]{8}{fig/comp_sota/pika/0267_}{0}{15} \\
    \rotatebox{90}{\normalsize VideoCrafter1} &
    \animategraphics[width=\linewidth]{8}{fig/comp_sota/videocrafter/0045_}{0}{15}&
    \animategraphics[width=\linewidth]{8}{fig/comp_sota/videocrafter/0330_}{0}{15}&
    \animategraphics[width=\linewidth]{8}{fig/comp_sota/videocrafter/0137_}{0}{15}&
    \animategraphics[width=\linewidth]{8}{fig/comp_sota/videocrafter/0267_}{0}{15} \\
    \rotatebox{90}{\normalsize Show-1} &
    \animategraphics[width=\linewidth]{8}{fig/comp_sota/show1/0045_}{0}{15}&
    \animategraphics[width=\linewidth]{8}{fig/comp_sota/show1/0330_}{0}{15}&
    \animategraphics[width=\linewidth]{8}{fig/comp_sota/show1/0137_}{0}{15}&
    \animategraphics[width=\linewidth]{8}{fig/comp_sota/show1/0267_}{0}{15} \\
    \rotatebox{90}{\normalsize AnimeDiff} &
    \animategraphics[width=\linewidth]{8}{fig/comp_sota/animatediff/0045_}{0}{15}&
    \animategraphics[width=\linewidth]{8}{fig/comp_sota/animatediff/0330_}{0}{15}&
    \animategraphics[width=\linewidth]{8}{fig/comp_sota/animatediff/0137_}{0}{15}&
    \animategraphics[width=\linewidth]{8}{fig/comp_sota/animatediff/0267_}{0}{15} \\
    \rotatebox{90}{\normalsize Ours} &
    \animategraphics[width=\linewidth]{8}{fig/comp_sota/ours/0045_}{0}{15}&
    \animategraphics[width=\linewidth]{8}{fig/comp_sota/ours/0330_}{0}{15}&
    \animategraphics[width=\linewidth]{8}{fig/comp_sota/ours/0137_}{0}{15}&
    \animategraphics[width=\linewidth]{8}{fig/comp_sota/ours/0267_}{0}{15} \\
  \end{tabular}
  \caption{Comparison of different text-to-video generation models. \textit{Best viewed with Acrobat Reader. Click the images to play the video clips.}}
  \label{fig:sota}
\end{figure*}

\fi

\subsection{Comparison with State-of-the-Art T2V Models}
We compare our approach with several state-of-the-art T2V models, including popular commercial models such as Gen-2~\cite{gen2} and Pika Labs~\cite{pikalab}, as well as open-source models like Show-1~\cite{zhang2023show1}, VideoCrafter1~\cite{chen2023videocrafter1}, and AnimateDiff~\cite{guo2023animatediff}.  
Gen-2, Pika Labs, and VideoCrafter1 all utilize high-quality videos for training their T2V models. 
It is noteworthy that AnimateDiff and our models use only the videos from WebVid-10M. 
Show-1 employs additional high-quality videos for finetuning to eliminate the watermark in WebVid-10M. 
AnimateDiff is not a generic T2V model; it works only when the LORA SD model is compatible with its temporal modules. 
For our comparison, we use its temporal modules (second version) based on SD v1.5 and employ Realistic Vision V2.0~\cite{realistic-vision-v20} as its corresponding LORA model.

\begin{table}[t] \footnotesize
{%
\begin{tabular}{>{\centering\arraybackslash}m{1.3cm}>{\centering\arraybackslash}m{1.2cm}>{\centering\arraybackslash}m{1.4cm}>{\centering\arraybackslash}m{1.2cm}>{\centering\arraybackslash}m{1.2cm}}
\toprule
{} & Visual & Text-Video & Motion & Temporal \\ 
{} & Quality & Alignment & Quality & Consistency \\ \midrule


 PikaLab$^*$   & 63.52   & 54.11   & 57.74  & 69.35    \\ 
 Gen2$^*$           & 67.35   & 52.30  & 62.53  & 69.71    \\ 

\hline
VideoCrafter1  & 61.64   &  66.76  & 56.06  & 60.36 \\
Show-1 & 52.19 & 62.07 & 53.74 & 60.83 \\
AnimeDiff & 58.89 & 74.79 & 51.38 & 56.61 \\
  Ours & 63.28 & 64.67 & 53.95 & 62.02\\
  
\bottomrule
\end{tabular}
}
\vspace{-1em}
\caption{
Comparison on the EvalCrafter benchmark. 
Higher score indicates better performance. 
* commercial models. 
}
\vspace{-1em}
\label{tb:sota}
\end{table}

\paragraph{Quantitative Evaluation.}
The quantitative results obtained using EvalCrafter are presented in Table~\ref{tb:sota}. 
Our method achieves visual quality comparable to that of VideoCrafter1 and Pika Labs, which use high-quality videos for training. 
This underscores the effectiveness of employing high-quality images to enhance picture quality and aesthetic scores. 
Furthermore, our text-video alignment performance is ranked second. 
In terms of motion quality, our performance surpasses that of Show-1 but falls short of models that utilize a larger volume of videos to learn motion. 
This indicates that our method can enhance visual quality without significant motion degradation.

\paragraph{Qualitative Evaluation.}
The visual comparison is depicted in Fig.~\ref{fig:sota}. Additional results are provided in the supplementary material. 
The visual quality of our results is on par with that of commercial models such as Gen-2 and Pika Labs. 
Since we employ JDB as the image dataset, the picture quality of our synthesized videos shifts from WebVid-10M to JDB. 
Regarding motion, our motion quality is superior to that of AnimateDiff and comparable to Show-1. 
Although the integration of temporal modules with a LORA SD model can enhance visual quality, AnimateDiff experiences motion degradation in generic scenes.

\begin{table}[t] \footnotesize
\centering

{%

\begin{tabular}{>{\centering\arraybackslash}m{2.8cm}
>{\centering\arraybackslash}m{1.4cm}
>{\centering\arraybackslash}m{1.2cm}
>{\centering\arraybackslash}m{1.2cm}} 
\toprule
Methods & Text-Video Alignment & Motion Quality & Visual Quality \\
\midrule
Ours vs Gen2 & 56\% & 46\% & 34\% \\
Ours vs AnimeDiff & 55\% & 64\% & 69\% \\
Ours vs Show-1 & 59\% & 59\% & 82\% \\
Ours vs VideoCrafter1 & 61\% & 63\% & 61\% \\
\bottomrule
\end{tabular}
}
\caption{
Human preference. The numbers represent the probability of users choosing our method. 
}
\label{tab:pertb-user}
\vspace{-0.05in}

\end{table}
\paragraph{User Study.}
For further evaluation, we conduct a user study to compare our method with other video models. 
We select $50$ prompts from EvalCrafter, covering diverse scenes, styles, and objects. 
When comparing a model pair, three video production experts are asked to select their preferred video from three options: method 1, method 2, and comparable results, according to the given subject, \textit{i.e.,} visual quality, motion quality, and text-video alignment. 
The results are shown in Table~\ref{tab:pertb-user}. 
Our method has better visual quality than AnimateDiff and Show-1 and is comparable to VideoCrafter1. 
Our method is more preferred than Show-1 and AnimateDiff in motion quality. 

\subsection{Strategy Evaluation}
\paragraph{Spatial-temporal Connection.}
In Sec.~\ref{sec:connection}, we show a visual comparison of perturbing the spatial and temporal parameters of the fully and partially trained models in Fig.~\ref{fig:connection-spatial} and Fig.~\ref{fig:connection-temp}. 
Here we provide the quantitative comparisons about the visual quality in Table~\ref{tab:pertb-vis}, including aesthetic and technical scores from DOVER~\cite{wu2023exploring}. 
We observe that finetuning the partially trained model can always achieve better visual quality than the fully trained model.
It means that the distribution of the partially trained model can be shifted more easily. 
Besides, we conduct a user study asking participants to choose a favorable model that has better performance in motion, in terms of foreground/background flash and motion flicker. 
The results are shown in Table~\ref{tab:pertb-user}. 
It can be observed that the motion quality of perturbed fully trained models is better. 
The fully trained model can tolerate larger parameter perturbations than the partially trained model. 
These observations show that the fully trained model has stronger spatial-temporal coupling. 

\begin{table}[t] \footnotesize
\centering

{%

\begin{tabular}{>{\centering\arraybackslash}m{2.4cm}>{\centering\arraybackslash}m{2.2cm}>{\centering\arraybackslash}m{2.2cm}} 
\toprule
Method & Aesthetic Score ($\uparrow$) & Technical Score ($\uparrow$) \\
\midrule
P-base & 34.32 & 42.69 \\
F-base & 46.55 & 51.76 \\
P-Spa-LORA & 78.25 & 72.74 \\
F-Spa-LORA & 77.97 & 59.60 \\
P-Temp-LORA & 77.40 & 54.85 \\
F-Temp-LORA & 66.26 & 50.32 \\
\midrule
F-Spa-DIR & 82.57 & 70.35 \\
F-Temp-DIR & 82.77 & 65.34 \\
F-Spa\&Temp-DIR & 83.59 & 67.75 \\
F-Spa\&Temp-LORA & 80.44 & 63.61 \\
F-Spa-DIR-LAION & 67.83 & 54.26 \\
\bottomrule
\end{tabular}
}
\caption{
Visual quality evaluation of the perturbed T2V models. 
}
\label{tab:pertb-vis}
\vspace{-0.05in}

\end{table}

\begin{table}[t] \footnotesize
\centering

{%

\begin{tabular}{>{\centering\arraybackslash}m{4.2cm} >{\centering\arraybackslash}m{3cm}} 
\toprule
Methods &  Motion Quality \\
\midrule
F-Spa-LORA vs P-Spa-LORA & 87\% \\
F-Temp-LORA vs P-Temp-LORA & 73\% \\
F-Spa-DIR vs F-Spa\&Temp-DIR & 67\% \\
\bottomrule
\end{tabular}
}
\caption{
User study on the motion of the perturbed T2V models. 
}
\label{tab:pertb-user}
\vspace{-0.05in}

\end{table}

\paragraph{Module Selection.}
After selecting the fully trained model as the base, we use two strategies to identify the most effective module to fine-tune, resulting in four models in Sec.~\ref{sec:disentanglement}.
The visual quality evaluation of these models is shown in the bottom part of Table~\ref{tab:pertb-vis}. 
The visual quality of F-Spa-DIR and F-Spa\&Temp-DIR is much better than the other two models. 
It reveals that directly finetuning spatial modules is the key to improving picture quality. 

Since F-Spa-DIR and F-Spa\&Temp-DIR achieve close visual quality, we conduct a user study on motion quality to determine the final model. The results are shown in the last row of Table~\ref{tab:pertb-user}. 
Directly finetuning the spatial modules only performs better in motion. 
As shown in Fig.~\ref{fig:disentangle-module}, F-Spa-DIR is more stable and has better temporal consistency than F-Spa\&Temp-DIR. 
The latter has obvious flashes in both the foreground and background. 

\paragraph{Influence of Image Data.}
To verify the effectiveness of synthesized images, we use the LAION Aesthetics V2 dataset and JDB to directly fine-tune the spatial modules in the second stage, respectively. 
The visual examples are shown in Fig.~\ref{fig:data}. 
It shows that the model trained with JDB composite concepts better than the model trained with LAION Aesthetics V2. 
The quantitative evaluation of visual quality is shown in Table~\ref{tab:pertb-vis}. 
F-Spa-DIR is much better than F-Spa-DIR-LAION in both aesthetic and technical scores.

\section{Conclusion}
\label{sec:conclusion}
To overcome data limitations, we propose a method for training high-quality video diffusion models without using high-quality videos. 
We delve into the training schemes of SD-based video models and investigate the coupling strength between spatial and temporal dimensions. 
We observe that fully trained T2V models exhibit stronger spatial-temporal coupling than partially trained models. 
Based on this observation, we propose disentangling appearance from motion at the data level, \textit{i.e.}, by exploiting low-quality videos for motion learning and high-quality images for appearance learning. 
Additionally, we suggest using synthetic images with complex concepts for finetuning, rather than real images. 
Quantitative and qualitative evaluations are conducted to demonstrate the effectiveness of the proposed method.

{\small
\bibliographystyle{ieeenat_fullname}
\bibliography{11_references}
}

\end{document}